# Investigating Deep Learning Approaches for Hate Speech Detection in Social Media

Prashant Kapil[1], Asif Ekbal[1], Dipankar Das[2]

[1] Indian Institute of Technology Patna, India
[1] Department of Computer Science and Engineering
[2] Jadavpur University Kolkata, India
[2] Department of Computer Science and Engineering
{prashant.pcs17,asif}@iitp.ac.in
ddas@cse.jdvu.ac.in

**Abstract.** The phenomenal growth on the internet has helped in empowering individual's expressions, but the misuse of freedom of expression has also led to the increase of various cyber crimes and anti-social activities. Hate speech is one such issue that needs to be addressed very seriously as otherwise, this could pose threats to the integrity of the social fabrics. In this paper, we proposed deep learning approaches utilizing various embeddings for detecting various types of hate speeches in social media. Detecting hate speech from a large volume of text, especially tweets which contains limited contextual information also poses several practical challenges. Moreover, the varieties in user-generated data and the presence of various forms of hate speech makes it very challenging to identify the degree and intention of the message. Our experiments on three publicly available datasets of different domains shows a significant improvement in accuracy and F1-score.

## 1 Introduction

Social media is one platform that allows people across the globe to share their views and sentiments on various topics, but when it is intended to hurt some particular group or any individual then it is considered as hateful content. There is no such universally accepted definition of hate speech as it often varies across the different geographical regions. [1] stated that hate speech is an abusive speech with a high frequency of stereotypical words. It is demographic dependent as some countries allow some speech to be said under *Right to speech*, whereas other countries adhere to a very strict policy for the same message.

In recent times, Germany made policy for the social media companies that they would have to face a penalty of 60$ million if they failed to remove illegal content on time. Denmark and Canada have laws that prohibit all the speeches that contain insulting or abusive content targeting minorities and could promote violence and social disorders. The Indian government has also urged leading social media sites such as Facebook, Twitter to take necessary action against hate speech, especially those posts that hurt

religious feelings and create social outrage. Setting aside legal actions our aim should be to combat these speeches by agreeing to a set of standard definitions, guidelines, and practices. [2] defined hate speech as any communication that demeans any person or any group based on race, color, gender, ethnicity, sexual orientation, and nationality. Social networking sites like Twitter and Facebook are also taking preventive measures by deploying hundreds to thousands of staff to monitor and remove offensive content.

[3] collected messages from Whisper and Twitter to define hate speech as any offense motivated, in whole or in a part, by the offender's bias against an aspect of a group of people. They investigated the main targets of hate speech in online social media and introduced new forms of hate that are not crimes but harmful.

The detection can't be done manually, rather it needs a thorough investigation of the techniques and build robust techniques to accomplish this task.
The paper is structured as follows: We put the discussion on the related works in Section 2. Section 3 describes embeddings used, preprocessing and the model architecture. Datasets and experimental setup are described in Section 4. Results along with the error analysis to discuss the limitations of our proposed models are presented in Section 5. Finally, we conclude along with future work roadmaps in Section 6.

### 1.1 Motivation and Contribution

There has not been much research on hate speech detection because of the non-availability of annotated datasets as well as lack of proper attention to this field. .

Its detection is challenging as these are highly contextual and poses several challenges concerning the demographic characteristics and nature of the text. The same message can be posted in different ways, with one could be the potential candidate for hate speech, while other is not. Data imbalance also introduces challenges to build a robust machine learning model. In this paper, we propose deep learning based approach to hate speech detection. We experimented with three publicly available benchmark datasets i.e [4], [5] and [6]

## 2    Related Work

Most of the previous works done in this area have used different data sets. Researchers have mostly used traditional machine learning algorithms, and recently have started using deep learning. Lexical based approaches misclassify any sentence containing slang indicative of hate, affecting *right to freedom of speech* as the word used may have different meaning used in some different contexts.
[7] showed that support vector machine (SVM) with word-n-grams employed with syntactic and semantic information can achieve the best performance. [5] reported that using unigram, bigrams, and trigrams feature weighted with their TF-IDF values fed to logistic regression(LR) tends to perform best on their dataset by achieving 90% precision with hate class correctly predicted for 61% times. [8] classified ontological classes of harmful speech based on the degree of content, intent, and affect that it is creating on social media. [4] used critical race theory to annotate a dataset of 16K tweets that is made publicly available. They observed that geographic and word length distributions

Investigating Deep Learning Approaches for Hate Speech Detection in Social Media    3do not have significant contributions in enhancing the performance of the classifier. However, gender information combined with char-n-grams has shown a little improvement.

[9] used four types of features like n-grams, linguistic features, syntactic features, and distributional semantic features to make a distinction between abusive and clean data in finance and news data. [10] made use of various semantic, sentiment and linguistic features to develop a cascaded ensemble learning classifier for identifying racist and radicalized intent on the Tumblr microblogging website.

[11] studied different forms of abusive behavior and made public the annotated corpus of 80K Tweets categorized into 8 labels. [12] classified 2010 sentences using features like unigrams, sentiment features, semantic features, and pattern-based features. [13] proposed a CNN-GRU based architecture that showed promising results for 6 out of 7 datasets, outperforming other state-of-the-art by 1-13 F1 points. They also released a new dataset of 2435 tweets focusing on refugees and Muslims.

[14] applied bag-of-words model to learn binary classifier for the labels *racist* and *non-racist* and achieved 76% accuracy. [15] used the combinations of neural network-based LSTM model with non-neural based GBDT representing words by random embedding and achieved the best result on the dataset of [4]. The method proposed in [16] focused on detecting abusive language first and then classify into specific types of abuse. They showed that hybrid CNN i.e a combination of char-cnn and word-cnn perform best over word-cnn and classical methods like logistic regression and svm on the dataset of 16K tweets by [4].

[17] showed the concept of using CNN with random vectors, word vectors based on semantic information, word vectors combined with character 4-grams, and compared the performance with each other.

## 3 Methodology

### 3.1 Pre-trained Word Vectors and One-Hot encoding

(a) *W2V*: We utilized the publicly available *word2vec* vectors trained on 100 billion words from Google news, trained using CBOW architecture [18] and have dimensions of 300.

(b) *GloVe*[19]: Training is performed on aggregated global word-word co-occurrence statistics from a corpus, and the resulting representations showcase interesting linear substructures of the word vector space. We used *glove.twitter.27B.100d* as the embeddings.
For (a) and (b) All the out-of-vocabulary (OOV) words were assigned random weights in the range [-0.25, 0.25].

(c) *FastText*: In our experiments, we also leveraged the skip-gram based approach by [20] that represents each word as a bag of character n-grams. A vector value is associated with each character, the sum of these vector values represent the embedding for words. The dimensions for these embedding are 300.

(d) *One-Hot encoding*: The encoding is done by prescribing an alphabet of size *m* for the input language, and then quantize each character using 1-of-m encoding. The alphabet used in all of our models consists of 27 characters, including 26 English letters and one for all other symbols.

### 3.2 Pre-processing

As the datasets have been crawled from social media, these contain noises and inconsistencies, such as slangs, misspelled words, acronyms, etc. Hence a light pre-processing is done by expanding all apostrophes containing words and then removing characters like : , & ! ?. The tokens were also converted to lower-case for normalization. We also used a dictionary to expand the misspelled words to its original form. All the words starting with # were broken down into individual words using word segment in python. For e.g. *#KillerBlondes* becomes *killer blondes*, *#Feminism* becomes *feminism*, *#atblackface* becomes *at black face* and *#marriageequality* becomes *marriage equality* etc. Emoticons were also replaced with tokens like happy, sad, disgust, and anger.

### 3.3 Models

We developed 13 deep learning models using CNN, LSTM, BiLSTM, and Character-CNN. Below we are describing the main models.

**CNN:** This model is based on the architecture by [21] that uses 5 main types of layers: *Input layer, Embedding Layer, Convolution layer, Pooling layer* and *Fully Connected layer*.

*Input Layer:* All the sequences are converted to integer form where each token has been assigned a unique index. The input sequences are then zero-padded to have an equal length as it helps in improving performance by keeping information preserved at the borders.

*Embedding Layer:* Each word $w_i$ in the sequence is mapped to real-valued vector at the corresponding index in the embedding matrix using e($w_i$), where *e* is the embedding matrix.

*Convolution Layer:* It is used to extract features for better representation of data using the learnable filter of size i*h, where *i* is the window size and *h* is the embedding dimension. Each filter is convolved through *i* words at a time and performs an element-wise dot product to get a feature $f_1$. This process is repeated (n-h+1) times to get the feature map F = [$f_1, f_2.....f_{n-h+1}$]. *N* number of filters are used to get the different feature maps.

*Pooling Layer:* It reduces the spatial size of the representation helping in reducing overfitting. Max pooling takes the local maximum value from the feature map depending on the pool size whereas global max pooling takes the pool size equal to the size of the input.

*Fully Connected layer:* The vectorized form of features obtained from the last CNN layer is fed into the fully connected layer which has every input connected to every output by weight. This is followed by the softmax activation function that calculates the probability values for all the classes. Fig.1 describes the sample architecture of CNN with dimension = 5.

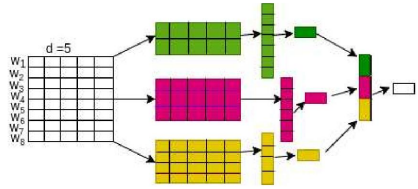

Fig. 1: **Architecture of CNN**

**LSTM/BiLSTM**: RNN is very suitable for sequence learning, time series but as it suffers from vanishing gradient and exploding gradient it does not perform well for the long-range dependency. So [22] introduced LSTM that is capable of learning long-range dependencies. The input sequence $(i_1,i_2...i_n)$ is transformed into its vector form of embedding size $e$ which is then converted to $h_1=(h_1^1,h_2^1...h_n^1)$ and transferred to the successive layers. It works by learning only the past information of the sequence, however, Bi-LSTM i.e a variant of LSTM comprises 2 LSTMs to capture both past and future information. At each time step the hidden state at any time sequence is the concatenation of forward and backward states $h_t=[\overrightarrow{h_t^1},\overleftarrow{h_t^1}]$, hence the input passed to next layer is $[e(w_1);h_1^1],[e(w_2);h_2^1],......,[e(w_n);h_n^1]$ as the input to the next layer is the concatenation of all the previous outputs. The next layer output will be $h_2 = (h_1^2,h_2^2....h_n^2)$. The input to the next layer will be $[e(w_1);h_1^1h_2^1,e(w_2);h_1^2h_2^2...]$. Fig. 2 shows the archititure of BiLSTM.

**Character-CNN**: We adopted the model of [23] that leverages the one-hot encoding to build the embedding matrix for the characters to represent sequences with 256 characters. Our designed model consists of representing each character using a 27 sized vector with 26 elements for the English alphabet and one for all other symbols. This model consists of a convolution layer with kernel size 4 followed by a max-pool layer of size 3. This is fed into another convolution layer with kernel size 4 and a max-pool layer of size 3. This is followed by 2 dense layers of size 64 and 2. The strides used in convolution layers are 4 and 2.

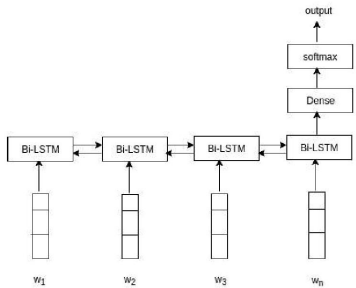

Fig. 2: **Architecture of BiLSTM**

## 4 Data sets

For the experiments, we use three types of datasets: **D1,D2** and **D3**. Table 1 shows the description of all the datasets with their total instances and the number of classes.
**D1**: This is the publicly available dataset with ≈ 16K Tweet IDs classified into three classes, *Racism*, *Sexism* and *Neither* by [4]. As some of the tweets were deleted as well as due to account suspension of the users we were able to retrieve around 15,476 tweets.

**D2**: This dataset is divided into three classes *Hate*, *Offensive* and *Neither* by [5].
**D3**: This is the aggressive data of English classified into *Overtly-Aggressive (OAG)*, *Covertly-Aggressive (CAG)* and *Non-Aggressive (NAG)* by [6].
**Table 2** shows the top occuring words in each sub-type of hate.

Table 1: **Details of the data set**

| Data | Total | Classes | #Tokens | Test Data |
|---|---|---|---|---|
| D1 | 15476 | Racism(1923)Sexism(2871) Neither(10682) | 12545 | CV* |
| D2 | 24783 | Hate(1430) Offensive(19190) Neither(4163) | 16362 | CV* |
| D3 | 15001 | OAG(3419)CAG(5297) NAG(6285) | 15830 | CV* $FB$: OAG(144), CAG(141), NAG(627) $SM$: OAG(361), CAG(413), NAG(483) |

*Test Data*:(*CV means there was no standard train/test split and thus 5-fold CV was used). *FB* is Facebook test data and *SM* is Social Media test data.

### 4.1 Experimental Setup

We use Keras [24] with Tensorflow [25] at the backend for our experiments. Experiments were performed using stratified 5-fold cross-validation to train all the classes

Investigating Deep Learning Approaches for Hate Speech Detection in Social Media     7Table 2: **List of top occuring words in each class**

| Class | High frequency words |
| --- | --- |
| Hate | b**ch, a**, nigger, f***ing, faggot, shit, trash, hate, kill, gay, ugly, queer, whitey |
| Offensive | b**ch, hoe, a**, ni**er, p***y, trash, wtf, crazy, stupid, p***s, gay, girl, hate |
| Racism | islam, religion, jews, women, war, christians, slave, terrorist, daesh, rape, beheaded |
| Sexism | sexist, women, girls, female, man, comedians, blondes, feminism, bitch, bimbos |
| Covertly | people, india, country, religious, party, political, muslims, fatwa, pakistan, modi, bjp |
| Overtly | people, india, religion, pakistan, bjp, muslims, hindu, terrorist, killed, fatwa |

according to their proportion. We report our results by accuracy and weighted F1-score. Categorical cross-entropy loss function and Adam optimizer were used for training because the former is very effective on the classification task than the classification error and mean square error [26]. Hidden nodes in LSTM and Bi-LSTM layers were set to 100. For regularization, dropout is applied to word embedding. The batch size of (16,32,64) and drop out of (0.1,0.2,0.3) were tested to build the model. The best accuracy and F1-score was obtained at 5 epochs and batch size of 32.

## 5 Results and Error Analysis

### 5.1 Results

Recurrent neural network based LSTM and BiLSTM performed best for all the 3 datasets. The addition of Char-CNN improved the overall accuracy and F-score. We are also discussing the existing approaches that were compared with our results in Table 3.

**Data 1**
[4]: Char n-grams obtained 73.89 weighted-F1 and char n-grams with gender information obtained 73.93 weighted-F1 using logistic regression classifier and 10-fold cross-validation.
[15]: Bag of words vectors(BoWV) uses the GloVe embedding with Gradient Boosted Decision Trees(GBDT), TF-IDF with GBDT and TF-IDF with SVM to obtain 80.10, 81.30 and 81.60 weighted-F1 by performing 10-fold cross-validation.

**Data 2**
[5]: Unigram, Bigrams, and Trigrams feature weighted by TF-IDF, Part-of-Speech tag unigram, bigrams, and trigrams fed into a logistic regression to obtain 90% weighted-F1.
[11]: They utilized text as well as a set of metadata features to obtain weighted-F1 of 89%.

**Data 3**
For Data 3 the results for Facebook(FB) test data and Social media(SM) test data were being reported by various teams participated in TRAC-1.
[27]: They developed LSTM and stacking of CNN-LSTM for Facebook and social media test data.
[28]: The TF-IDF and latent semantic analysis (LSA) were computed for character and word n-gram features.

8       Prashant Kapil[1], Asif Ekbal[1], Dipankar Das[2]

Table 3: **Results**

| Model | D1 | | D2 | | D3 | |
|---|---|---|---|---|---|---|
| | Accuracy | F1-Score | Accuracy | F1-Score | Accuracy | F1-Score |
| | | | | | CV/FB/SM | CV/FB/SM |
| 1. CNN(W2V) | 90.47 | 89.81 | 83.15 | 82.67 | 56.63/60.63/61.49 | 56.04/**63.14**/56.01 |
| 2. LSTM(W2V) | 90.55 | 89.51 | 83.57 | 83.24 | 57.26/58.55/61.41 | 57.17/62.18/59.33 |
| 3. BILSTM(W2V) | 90.95 | 89.36 | 83.98 | 83.53 | 58.45/55.70/58.47 | 58.30/59.45/58.57 |
| 4. CNN(Glove) | 90.35 | 89.60 | 82.98 | 82.61 | 55.97/60.19/61.65 | 55.31/62.47/59.19 |
| 5. LSTM(Glove) | 91.07 | 89.48 | 84.14 | 83.88 | 57.50/57.67/**64.20** | 57.53/61.57/62.26 |
| 6. BILSTM(Glove) | 91.08 | 89.99 | 84.25 | **83.95** | 58.48/55.26/59.90 | 58.30/59.07/59.07 |
| 7. CNN(Fasttext) | 90.17 | 88.85 | 83.21 | 82.66 | 56.06/55.26/61.49 | 55.17/58.64/57.40 |
| 8. LSTM(Fasttext) | 90.66 | 88.65 | 83.77 | 83.44 | 57.26/54.82/63.56 | 57.37/58.72/**62.67** |
| 9. BiLSTM(Fasttext) | 91.08 | 89.67 | 84.13 | 83.82 | 58.20/55.70/58.47 | 58.09/59.21/58.57 |
| 10. CharCNN | 87.34 | 85.12 | 79.98 | 78.55 | 46.55/53.83/44.15 | 43.07/55.01/42.03 |
| 11. LSTM(Glove)+CharCNN | 90.63 | 89.04 | 83.93 | 83.69 | 57.28/58.66/57.27 | 56.85/61.82/57.52 |
| 12. BiLSTM(Glove)+CharCNN | **91.09** | **90.39** | 84.14 | 83.88 | **58.83**/59.64/61.33 | **58.72**/62.83/59.88 |
| 13. BiLSTM(Fasttext)+CharCNN | 90.67 | 89.34 | **85.86** | 82.61 | 57.37/**60.74**/61.25 | 57.22/63.11/61.49 |
| Existing State of the Art | | | | | | |
| [4] | - | - | - | 73.89 | - | - |
| [4] | - | - | - | 73.93 | - | - |
| [15] | - | - | - | 80.10 | - | - |
| [15] | - | - | - | 81.30 | - | - |
| [15] | - | - | - | **81.60** | - | - |
| [5] | - | 90.00 | - | - | - | - |
| [11] | - | 89.00 | - | - | - | - |
| [27] | - | - | - | - | —/**62.28**/61.73 | —/**64.25**/59.20 |
| [28] | - | - | - | - | —/60.96/59.02 | —/63.15/57.16 |
| [29] | - | - | - | - | —/58.44/59.10 | —/61.78/55.20 |
| [30] | - | - | - | - | —/58.22/57.43 | —/61.60/56.50 |
| [31] | - | - | - | - | —/56.47/60.86 | —/60.11/59.95 |
| [32] | - | - | - | - | —/54.71/60.14 | —/58.13/**60.09** |

[29]: They utilized LSTM and CNN leveraging fasttext for Facebook and social media data.
[30]: SVM and BiLSTM model obtained best results for twitter and Facebook data.
[31]: They combined Gated recurrent unit (GRU) with three logistic regression classifiers trained on character, word n-grams, and hand-picked syntactic features.
[32]: The designed model with a Dense architecture performs better than a Fasttext model for both social media and Facebook data.

### 5.2 Error Analysis

Error analysis was carried out to analyze the errors that were encountered in our system So we analyzed the best model confusion matrix as they were giving better performance. We did the quantitative analysis in terms of the confusion matrix and qualitative analysis for analyzing the misclassified tweets.
**Quantitative analysis**: Table 4 enlists the confusion matrix for Data 1 and Data 2 obtained by BiLSTM(Glove) and BiLSTM(Glove) concatenated with Character-CNN. Ta-ble 5 consists of a confusion matrix obtained by training Data 3 in cross-validation utilizing Model 12. It also contains the confusion matrix generated by testing the model with social media and facebook test data. From Table 6 we can infer that identifying Hate, Overtly, and Covertly classes posses more challenges than other subclasses. Apart from data imbalance, using the sarcastic phrase and racial epithets in a deceitful manner makes it challenging for the classifier to identify hate sentences that had 70.97% false-negative rate and with only 29.02% true positive in D2. Due to some common obscene words between hate and offensive classes, 1.74% of offensive instances converted to hate.

Table 4: **Confusion Matrix for D1(Model 6) and D2(Model 12)**

| Class | Dataset 1 | | | Class | Dataset 2 | | |
|---|---|---|---|---|---|---|---|
| | Racism | Sexism | Neither | | Hate | Offensive | Neither |
| Racism | 1538 | 14 | 371 | Hate | 415 | 861 | 154 |
| Sexism | 17 | 1800 | 1054 | Offensive | 334 | 18347 | 509 |
| Neither | 539 | 441 | 9702 | Neither | 43 | 306 | 3814 |

Table 5: **Confusion Matrix for D3(Model 12) and D3(FB(Model 1) and SM(Model 8))**

| Class | Dataset 3(CV) | | | Class | Dataset 3(FB and SM) | | |
|---|---|---|---|---|---|---|---|
| | OAG | CAG | NAG | | FB/SM OAG | FB/SM CAG | FB/SM NAG |
| OAG | 1601 | 1285 | 533 | OAG | 77/237 | 35/108 | 32/16 |
| CAG | 921 | 2842 | 1534 | CAG | 32/147 | 50/160 | 59/106 |
| NAG | 371 | 1531 | 4383 | NAG | 63/14 | 138/67 | 426/402 |

Table 6: **Metric Values**

| Class | True Positive | False Positive | False Negative |
|---|---|---|---|
| Racism | 79.97 | 26.58 | 20.02 |
| Sexism | 62.69 | 20.17 | 37.30 |
| Hate | 29.02 | 47.60 | 70.97 |
| Offensive | 95.60 | 5.98 | 4.40 |
| Overtly | 46.82 | 44.65 | 53.18 |
| Covertly | 53.65 | 49.77 | 46.34 |

**Qualitative analysis**: For each data set we perform qualitative analysis to analyze the errors and we find that due to hate language being contextual in nature and also when the attack is directly or indirectly on women, then the model is showing poor performance. This suggests that it is indeed difficult for models to classify into fine-grained labels. Table 7 contains some of the sentences converted to different classes due to system inefficiency.

### 5.3 Statistical Significance Test

We also determine whether a difference between the worst and the best classifier i.e Character-CNN and BiLSTM(GloVe) + Character-CNN is statistically significant (at p $\leq 0.05$), for this we run a bootstrap sampling test on the predictions of two systems. The test takes 3 confusion matrix out of 5 at a time and compares whether the better system is the same as the better system on the entire data set. The resulting (p-) value of the bootstrap test is thus the fraction of samples where the winner differs from the entire data set. Table 8 depicts the statistical significance test performed on all 3 data sets.

Table 7: **Example of a sentence predicted to different class**

| Data | Original | Predicted | Tweet |
| --- | --- | --- | --- |
| D1 | Sexism | Neither | Please women stay single please women when you commit to your man,commit to the gym as well. |
| D1 | Racism | Neither | @AdnanSadiq01 I think your goat is calling you. She is horny.. |
| D1 | Sexism | Racism | hate watch you have to be extra stupid to be a women and follow #Islam. |
| D1 | Neither | Sexism | As long as she realizes she's not gonna look as pretty as she usually works.This character is a kind of mess. |
| D2 | Hate | Offensive | @Fit4LifeMike @chanelisabeth hoe don't make me put up screenshots of your texts to me hoe. |
| D2 | Hate | Offensive | @vinny2vicious faggot I knew you weren't really my friend. |
| D2 | Hate | Neither | They should have never gave a cracker a transmitter!!!!!! @realdjTV will flip when he sees this. |
| D3 | Covertly | Overtly | I told you wait.7 pak killed within hours of their cowardice act.Go and weep for them. |
| D3 | Overtly | Covertly | yes we remember you are biggest terrorist country in the world you will do anything against humanity. |

Table 8: **Bootstrapping Test**

| Data | Total | Sample taken | p-value |
| --- | --- | --- | --- |
| D1 | 15476 | 60% | $\leq 0.01$ |
| D2 | 24783 | 60% | $\leq 0.01$ |
| D3 | 15001 | 60% | $\leq 0.07$ |

## 6  Conclusion and Future work

In this paper we have explored the effectiveness of deep neural network for hate speech detection. The system failure on some cases highlights the subjective biases while classifying gender based message. Transfer learrning using large datasets can be very effective. Also some other linguistic features focused on gender and location will be used to improve the performance of the system. Some more other forms of hate will also be considered.

## Acknowledgement

The first author would like to acknowledge the funding agency, the University Grant Commission (UGC) of the Government of India, for providing financial support in the form of UGC NET-JRF.